\documentclass{article}
\usepackage{spconf,amsmath,graphicx}
\usepackage{multirow}
\usepackage{subcaption}
\usepackage{booktabs}
\usepackage{amsfonts}
\usepackage{array}
\makeatletter
\usepackage{arydshln}

\newcommand{\thickhline}{%
    \noalign {\ifnum 0=`}\fi \hrule height 1pt
    \futurelet \reserved@a \@xhline
}
\newcommand{\thinhline}{%
    \noalign {\ifnum 0=`}\fi \hrule height 0.25pt
    \futurelet \reserved@a \@xhline
}
\newcolumntype{"}{@{\hskip\tabcolsep\vrule width 1pt\hskip\tabcolsep}}

\title{Rethinking Efficacy of Softmax \\ for Lightweight Non-Local Neural Networks}
\name{Yooshin Cho$^{1}$, Youngsoo Kim$^{1}$, Hanbyel Cho$^{1}$, Jaesung Ahn$^{2}$, Hyeong Gwon Hong$^{2}$, and Junmo Kim$^{1,2}$}
\address{$^{1}$School of Electrical Engineering, KAIST, South Korea \\
$^{2}$Kim Jaechul Graduate School of AI, KAIST, South Korea}
\begin{document}
%
\maketitle
\begin{abstract}
Non-local (NL) block is a popular module that demonstrates the capability to model global contexts. However, NL block generally has heavy computation and memory costs, so it is impractical to apply the block to high-resolution feature maps. In this paper, to investigate the efficacy of NL block, we empirically analyze if the magnitude and direction of input feature vectors properly affect the attention between vectors. The results show the inefficacy of \textit{softmax} operation which is generally used to normalize the attention map of the NL block. Attention maps normalized with \textit{softmax} operation highly rely upon magnitude of key vectors, and performance is degenerated if the magnitude information is removed. By replacing \textit{softmax} operation with the scaling factor, we demonstrate improved performance on CIFAR-10, CIFAR-100, and Tiny-ImageNet. In Addition, our method shows robustness to embedding channel reduction and embedding weight initialization. Notably, our method makes multi-head attention employable without additional computational cost.
\end{abstract}
\begin{keywords}
Attention, Non-local block, Transformer
\end{keywords}

\vspace{-1mm}
\section{Introduction}
\label{sec:intro}
\vspace{-2mm}
Self-attention layers such as Non-Local (NL) block~\cite{wang2018non} and Transformer~\cite{vaswani2017attention} were proposed to capture long-term dependencies, and considered as a key component in Natural Language Process (NLP) deep learning architectures~\cite{devlin2018bert, dai2019transformer, wang2020linformer, lan2019albert}. To capture global features, self-attention layers model relationship between pixels regardless of distance. This property benefits not only machine translation, but also most computer vision tasks. However, NL blocks have been employed in a limited manner in computer vision owing to their heavy computation and memory cost that increases as a quadratic function of the number of pixels. Generally the number of pixels is much larger than the number of words, and thus the cost is not scalable to realistic input image sizes.

Evidently, reducing the cost of NL blocks is still an active research area~\cite{huang2019ccnet, li2020neural, vaswani2021scaling, dai2019transformer, levi2018efficient, lu2021soft, richter2020normalized}. Previous studies have focused on introducing lightweight NL blocks and methods to efficiently employ NL blocks. They suggested lightweight NL blocks by efficiently reducing spatial size~\cite{huang2019ccnet, vaswani2021scaling, dosovitskiy2020image} and approximating the attention while minimizing the loss of the capability to capture long-term dependencies. To optimize the trade-off between the capability to obtain global relationships and computational efficiency, previous methods have relied on heuristic, adopted approximation or neural architecture search (NAS) algorithms~\cite{stamoulis2019single, tan2019mnasnet, li2020neural}. These methods demonstrated plausible performance and reduced computational overhead, but reducing the spatial size and the number of NL block cannot avoid the loss of the capability to incorporate global context.

\begin{figure*}
\centering
\begin{tabular}{ccc}
\includegraphics[width=3.2cm]{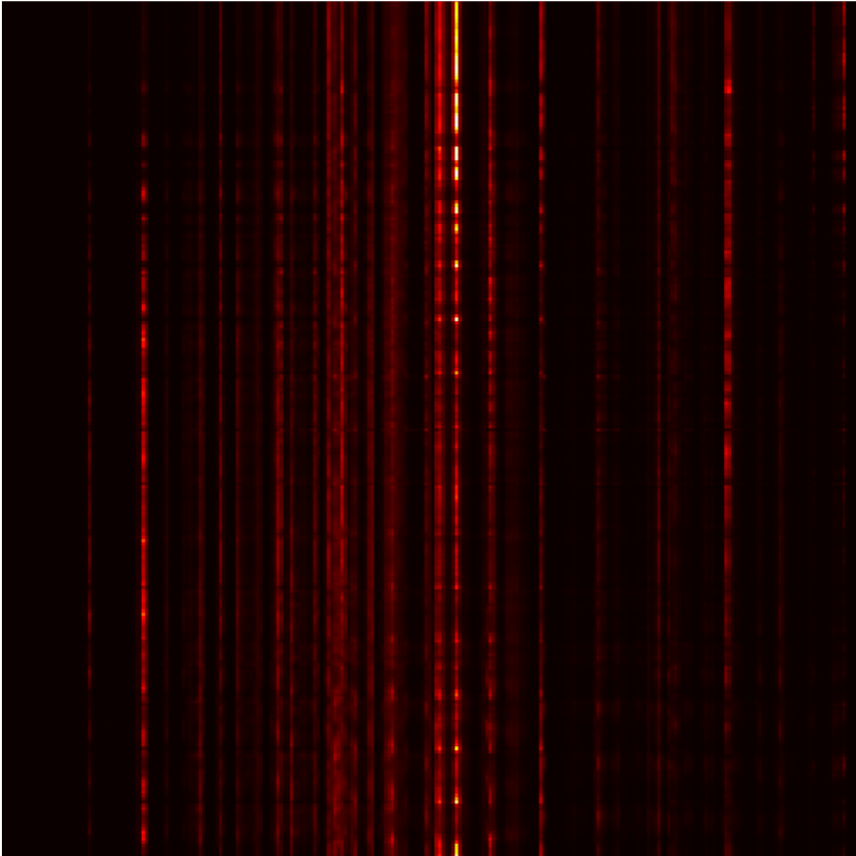}&
\includegraphics[width=3.2cm]{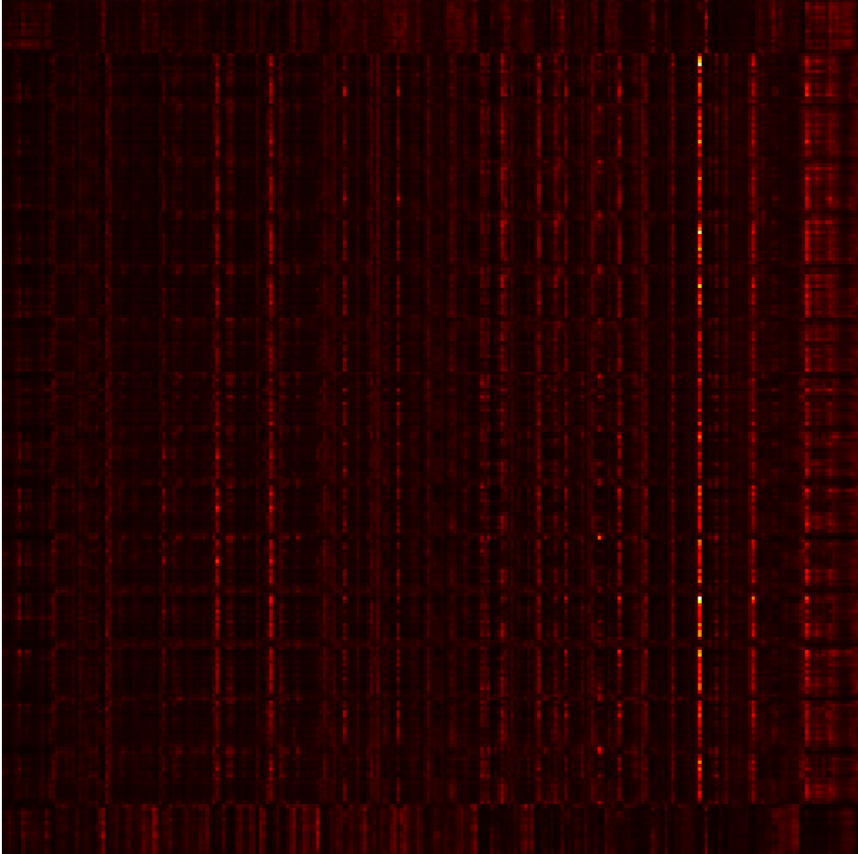}&
\includegraphics[width=3.2cm]{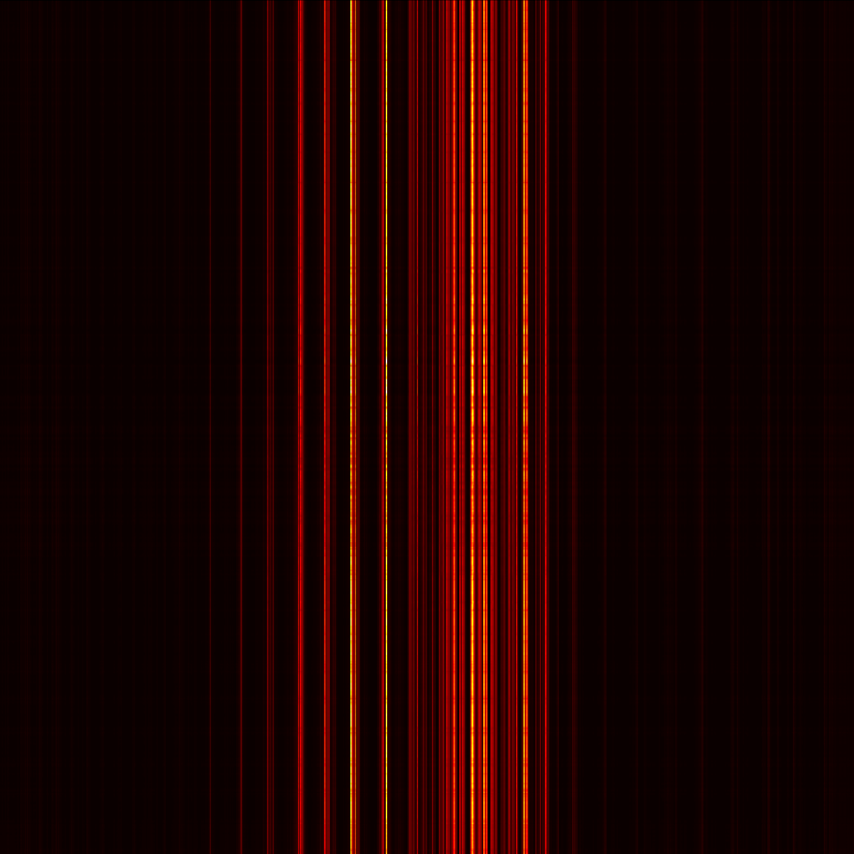}
\\
(a) CIFAR-10 / NL &(b) CIFAR-100 / NL &(c) Tiny-ImageNet / NL
\\
\includegraphics[width=3.2cm]{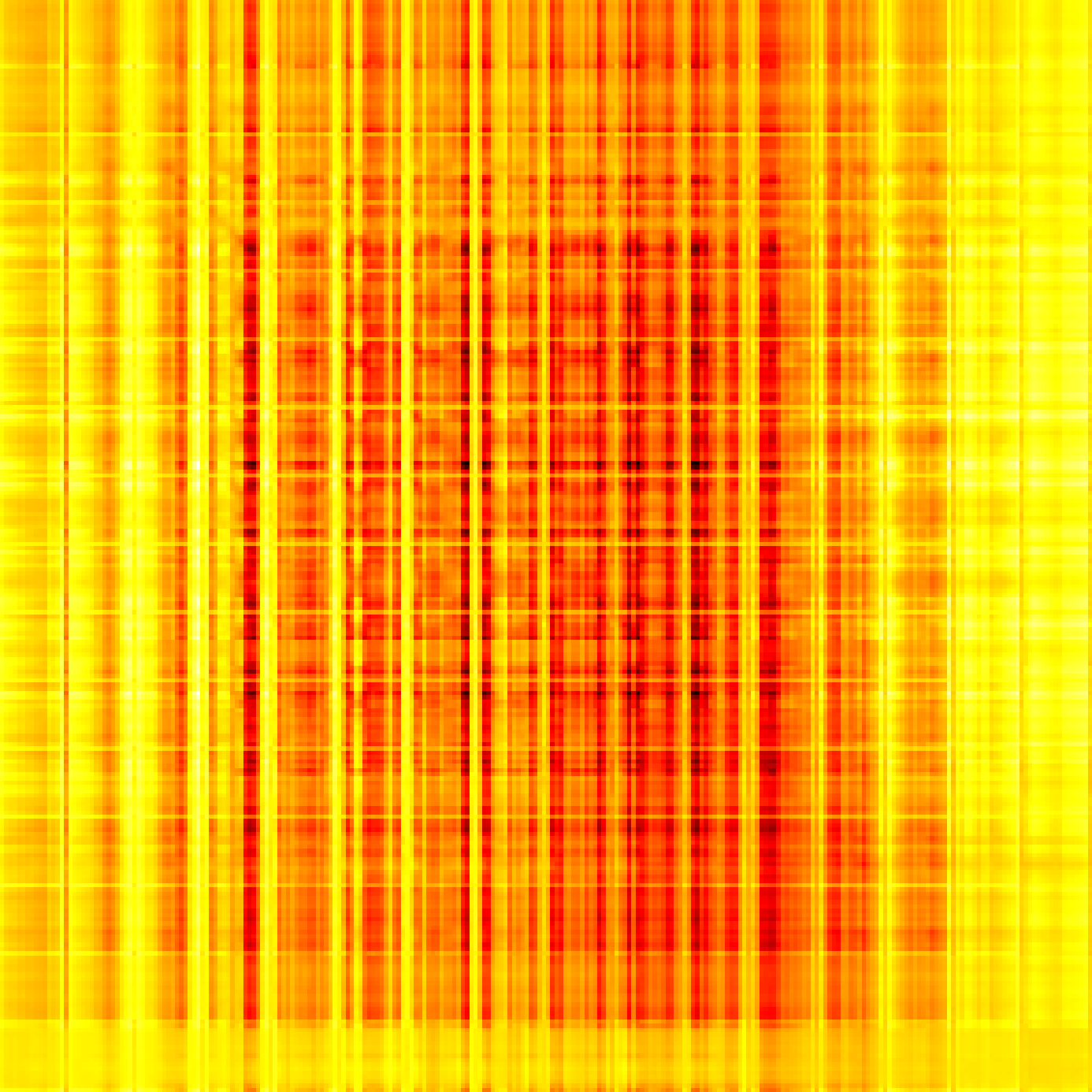}&
\includegraphics[width=3.2cm]{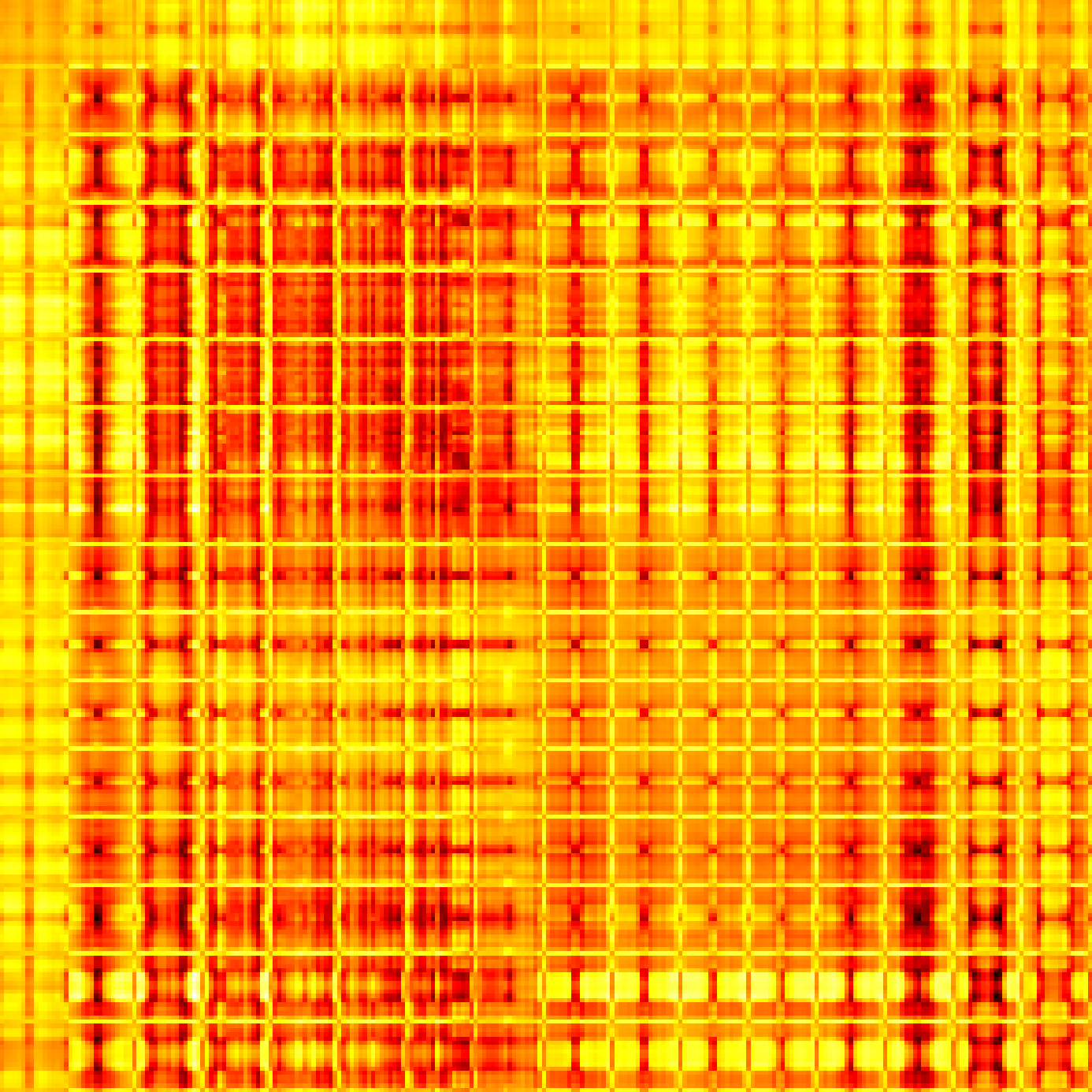}&
\includegraphics[width=3.2cm]{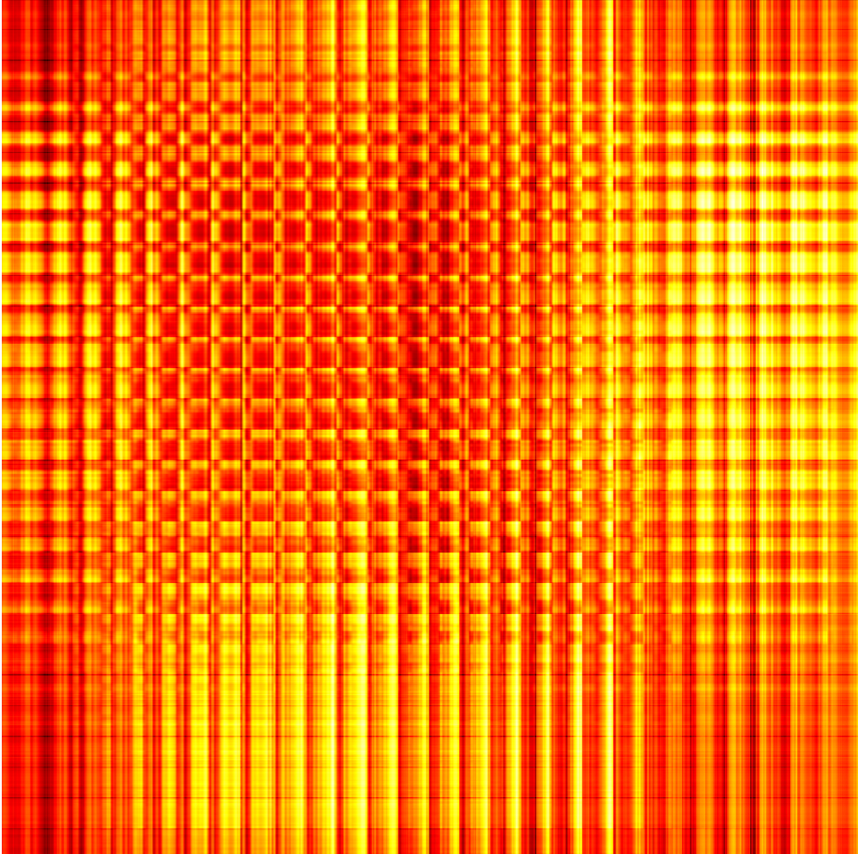}
\\
(d) CIFAR-10 / Ours &(e) CIFAR-100 / Ours&(f) Tiny-ImageNet / Ours
\end{tabular}
\vspace{-3mm}
\caption{Visualization of the attention maps of randomly sampled images. Attention maps are size of $HW \times HW$, where the X-axis and Y-axis are key and query pixels, respectively. Attention maps of NL block are clearly affected by few key pixels, and vertical lines are observed. By contrast, attention maps of our method does not demonstrate a vertical line.}
\vspace{-3mm}
\label{fig:att map}
\end{figure*}

In this paper, we empirically analyze the efficacy of \textit{softmax} operation of NL blocks using the geometric definition of the dot product. In most cases, attention is computed using the dot-product and normalized with \textit{softmax} operation~\cite{wang2018non, vaswani2017attention, dai2019transformer, huang2019ccnet, vaswani2021scaling, devlin2018bert, dosovitskiy2020image, lan2019albert}. Geometrically, dot-product is a multiplication of magnitudes and cosine similarity between two pixel vectors. From this perspective, we suspect that \textit{softmax} operation makes modeling relationship using cosine similarity inefficient for the following reason. To focus on angular relationships, let's assume that query and key vectors have a unit norm. Then, if \textit{softmax} operation is employed to normalize attention map, attention between query and key is minimized when dot product is $-1$. Thus, for a single key vector to have the minimum attention with more than two queries, those queries should have the same direction, reducing angular variation of queries. However, if attention is not normalized by \textit{softmax} operation, extremely low attention can be expressed by orthogonality between queries and keys. For a single key vector, queries with zero attention can be diversely selected in its hyperplane of dimension $C-1$ that is orthogonal to the key vector, where $C$ is the channel size of queries and keys. Hence, we suspect \textit{softmax} operation might limit the capability to model relationships, and make NL block dependent on magnitude rather than direction of vectors.

To verify our assumption, we train PreResNet~\cite{he2016identity} with NL blocks on CIFAR-10/100 and Tiny-ImageNet~\cite{krizhevsky2009learning,le2015tiny}, and visualize the attention maps of NL block with randomly sampled images in Figure~\ref{fig:att map}. Attention maps are matrices of size $HW \times HW$ which are computed by Eq~\ref{eq:att}. Attention maps of NL block demonstrate clear vertical lines; it indicates that attention value rarely changes despite varying query, and attention is dominantly affected by keys itself (e.g., magnitude). In other word, cosine similarity is less discriminatively trained as we suspected. Therefore, we introduce the \textit{scaled NL block} that does not employ \textit{softmax} operation, and it will be described in Section~\ref{subsec:ScaledNL}. As illustrated in Figure~\ref{fig:att map}, our method does not demonstrate a straight vertical line, and it indicates that attention depends on both queries and keys.

\textit{Scaled NL block} shows two beneficial properties owing to our method properly utilizes the direction of feature vectors. First, \textit{scaled NL block} shows robustness to embedding channel reduction. Because the proposed method efficiently utilizes the embedding feature space, performance degradation due to embedding channel reduction is significantly smaller than for the NL block. Second, \textit{scaled NL block} demonstrates robustness to embedding weight initialization. NL block performs better when the weight of the embedding layer is initialized with a standard deviation of $0.01$, which is tuned hyper-parameter. By contrast, proposed method is suitable to \textit{He initialization}~\cite{he2015delving} which is the standard initialization method. In addition, we generally obtain better performance on PreResNet~\cite{he2016identity} and Wide-Residual Network (WRN) ~\cite{zagoruyko2016wide} with CIFAR-10/100~\cite{krizhevsky2009learning}, and Tiny ImageNet~\cite{le2015tiny}. Finally, we investigate the memory consumption and train step time of multi-head attention of NL blocks. The memory consumption and train step time of NL block are linear functions of the number of heads. By contrast, our method makes multi-head attention adoptable without additional computation cost.

\vspace{-2mm}
\section{Related Works}
\label{sec:related}
\vspace{-2mm}
The monumental attention layer, Transformer~\cite{vaswani2017attention} achieved the best performance at the time on machine translation tasks based solely on attention mechanisms. Wang~\textit{et al.}\cite{wang2018non} employed attention mechanisms in computer vision applications to incorporate the global spatio-temporal context. Since their success, these self-attention layers have been widely used to model long-range relationships in variou applications~\cite{dai2019transformer,devlin2018bert, huang2019ccnet,levi2018efficient, li2020neural, vaswani2021scaling, dosovitskiy2020image}. Self-attention layers can be expressed generally using the following formula:
\vspace{-2mm}
\begin{equation}
\label{eq:att}
    \boldsymbol{A_{i,j}} = \frac{1}{Z(\textbf{x})}f(\boldsymbol{x_i},\boldsymbol{x_j}),
    \vspace{-2mm}
\end{equation}
\vspace{-1mm}
\begin{equation}
\label{eq:NL}
    \boldsymbol{y_{i}} =\sum_{\forall j} \boldsymbol{A_{i,j}}g(\boldsymbol{x_j}),
    \vspace{-2mm}
\end{equation}
where $\boldsymbol{x},\boldsymbol{y} \in \mathbb{R}^{HW \times C}$ are the input and output of the NL block, and $i, j$ are the indices of the query and key pixels. $\boldsymbol{A} \in \mathbb{R}^{HW \times HW}$ is the attention map. In most cases, NL block takes the form $f(\boldsymbol{x_i}, \boldsymbol{x_j})=e^{\frac{1}{\sqrt{C}}\theta (\boldsymbol{x_i}) \cdot \phi (\boldsymbol{x_j})}$ and $Z(\boldsymbol{x})=\sum_{\forall j} f(\boldsymbol{x_i}, \boldsymbol{x_j})$, where $\theta, \phi, g$ are the linear embedding layers. In this case, NL block gets attention from the embedded dot product that normalized with \textit{softmax} operation.

\section{Geometrical Analysis}
\label{sec:Geomatric}
\vspace{-2mm}
\subsection{Scaled Non-Local Block}
\label{subsec:ScaledNL}
\vspace{-1mm}
Previous studies~\cite{levi2018efficient,li2020neural,lu2021soft,richter2020normalized} have suggested attention modules without \textit{softmax} operation for improved computational efficiency. However, the inefficacy of \textit{softmax} operation has not been fully analyzed yet. To empirically analyze the inefficacy of \textit{softmax} operation, we introduce the NL block without \textit{softmax} operation. Instead of \textit{softmax} operation, we divide the output of NL block by $\sqrt{HW}$ to stabilize the block, where $H,W$ are the height and width of the input matrix, respectively\footnote{Assume that elements of $g(x)$ and $A=\frac{1}{\sqrt{C}}\theta(\boldsymbol{x})\cdot \phi(\boldsymbol{x})^{\top}$ are independent gaussian random variables with mean 0 and variance 1. Then, $A \cdot g(x)$ has mean 0 and variance $HW$. Thus, we scaled the output by $\sqrt{HW}$.}. We empirically verify that the proposed method without scaling factor is often diverging, but scaling successfully prevents divergence. In this paper, we denote the block as \textit{scaled Non-local Block}. If \textit{softmax} operation is replaced with the scaling factor, Eq~\ref{eq:NL} can be expressed as follows by employing the associative law:
\vspace{-2mm}
\begin{equation}
\label{eq:NL_assoc}
\begin{split}
    \boldsymbol{y} &= \frac{1}{\sqrt{HW}}(\frac{1}{\sqrt{C}}\theta(\boldsymbol{x})\cdot \phi(\boldsymbol{x})^{\top}) \cdot g(\boldsymbol{x}) \\
    &=  \frac{1}{\sqrt{HWC}}\theta(\boldsymbol{x})\cdot (\phi(\boldsymbol{x})^{\top} \cdot g(\boldsymbol{x})),
    \vspace{-2mm}
\end{split}
\end{equation}
where $\theta$, $\phi$, and $g$ are linear embedding layers. As suggested in~\cite{levi2018efficient,li2020neural}, it can largely reduces the computational cost by employing the associative law, even two forms are numerically equivalent. In the following sections, we compare the properties of NL block and \textit{scaled NL block} to demonstrate the inefficacy of \textit{softmax} operation. 

\vspace{-1mm}
\subsection{Importance Analysis}
\label{subsec:ImpAnls}
\vspace{-1mm}

As mentioned earlier, we suspect that \textit{softmax} operation limits the capability to model relationships between vectors, because it reduces the angular variations of query vectors having zero attention to a single key vector. For this reason, we assume that the cosine similarity terms of the dot-product is inefficiently learned, and attention maps highly rely on the magnitude of key vectors. To verify our assumption, we illustrate attention maps of the NL block in Figure~\ref{fig:att map}, which demonstrate clear vertical lines. This indicates that attention map are highly affected by the magnitude of key vectors. By contrast, attention maps of \textit{scaled NL block} do not show vertical lines. For further investigation, we train PreResNet with a \textit{magnitude only NL block} and a \textit{direction only NL block}, which are respectively expressed by the following formulas: 
\vspace{-1mm}
\begin{equation}
        \theta_{mag}(\boldsymbol{x_i}) = \lVert \theta(\boldsymbol{x_{i}}) \rVert  \text{,} \quad
        \phi_{mag}(\boldsymbol{x_i}) = \lVert \phi(\boldsymbol{x_{i}}) \rVert \text{,}
\end{equation}
\vspace{-2mm}
\begin{equation}
        \theta_{dir}(\boldsymbol{x_i}) = \frac{\theta(\boldsymbol{x_{i}})}{\lVert \theta(\boldsymbol{x_{i}}) \rVert} \text{,} \quad
        \phi_{dir}(\boldsymbol{x_i}) = \frac{\phi(\boldsymbol{x_{i}})}{\lVert \phi(\boldsymbol{x_{i}}) \rVert} \text{.}
\end{equation}
To verify whether NL blocks properly utilize the cosine similarity information, we replace the $\{\theta, \phi\}$ of Eq~\ref{eq:NL} with $\{\theta_{mag}, \phi_{mag}\}$ or $\{\theta_{dir}, \phi_{dir}\}$. As shown in Table~\ref{tab:mag,angle}, the performance of \textit{direction only NL block} is severely worse than for the \textit{direction only scaled NL block}. By constrast, \textit{magnitude only scaled NL block} demonstrates comparable performance with \textit{magnitude only NL block}. This indicates that by replacing $softmax$ operation to scaling factor, the capability to utilize angular information is improved while the capability to utilize magnitude information is maintained.

\begin{table}
\begin{center}
\tabcolsep=2mm
\resizebox{0.47\textwidth}{!}
{
\begin{tabular}{c|c|c|c|c}
\thickhline
  Dataset & Model & Base & Mag & Dir\\
\thickhline
\multirow{2}{*}{CIFAR-10}&PreResNet56+3NL & \textbf{5.73} & \underline{5.83} & 6.01  \\
&PreResNet56+3Ours & \textbf{5.64} & 5.76 & \underline{5.67} \\
\hline
\multirow{2}{*}{CIFAR-100}&PreResNet56+3NL & \textbf{25.12} & \underline{25.26} & 25.44  \\
&PreResNet56+3Ours & \textbf{24.53} & 25.20 & \underline{24.68} \\
\thickhline
\end{tabular}
}
\end{center}
\vspace{-3mm}
\caption{ Comparison of test errors($\%$) on PreResNet56 with CIFAR-10/100. Results are averaged over 10 random seeds. Base refers to NL block utilizing both magnitude and direction of vectors.}
\vspace{-2mm}
\label{tab:mag,angle}
\end{table}

\subsection{Robustness}
\label{subsec:robust}
\vspace{-1mm}

\begin{figure}[t]
    \centering
    \begin{subfigure}[b]{0.23\textwidth}
        \centering
        \includegraphics[width=\textwidth]{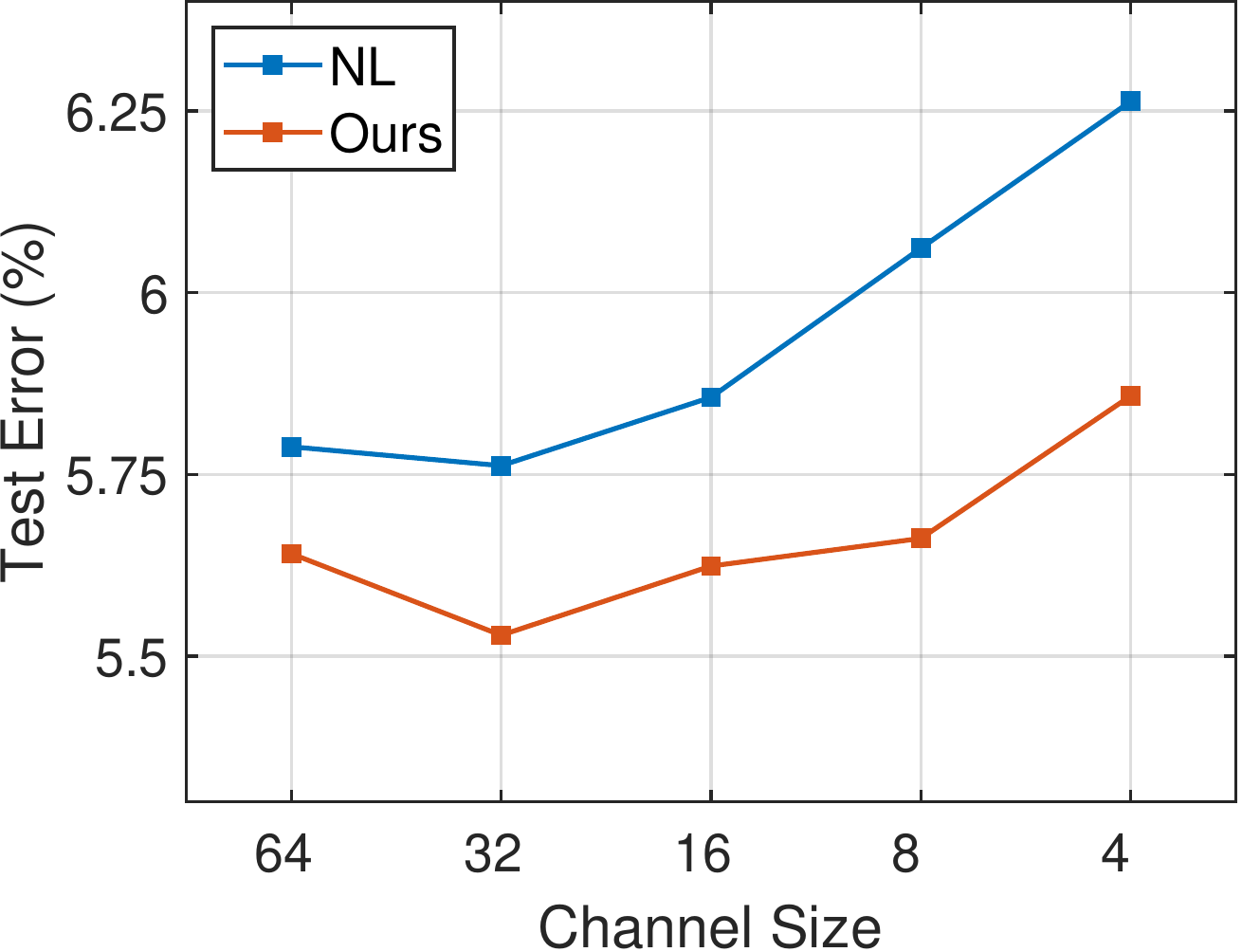}
        \subcaption{CIFAR-10}
    \end{subfigure}
    \hfill
    \begin{subfigure}[b]{0.23\textwidth}
        \centering
        \includegraphics[width=\textwidth]{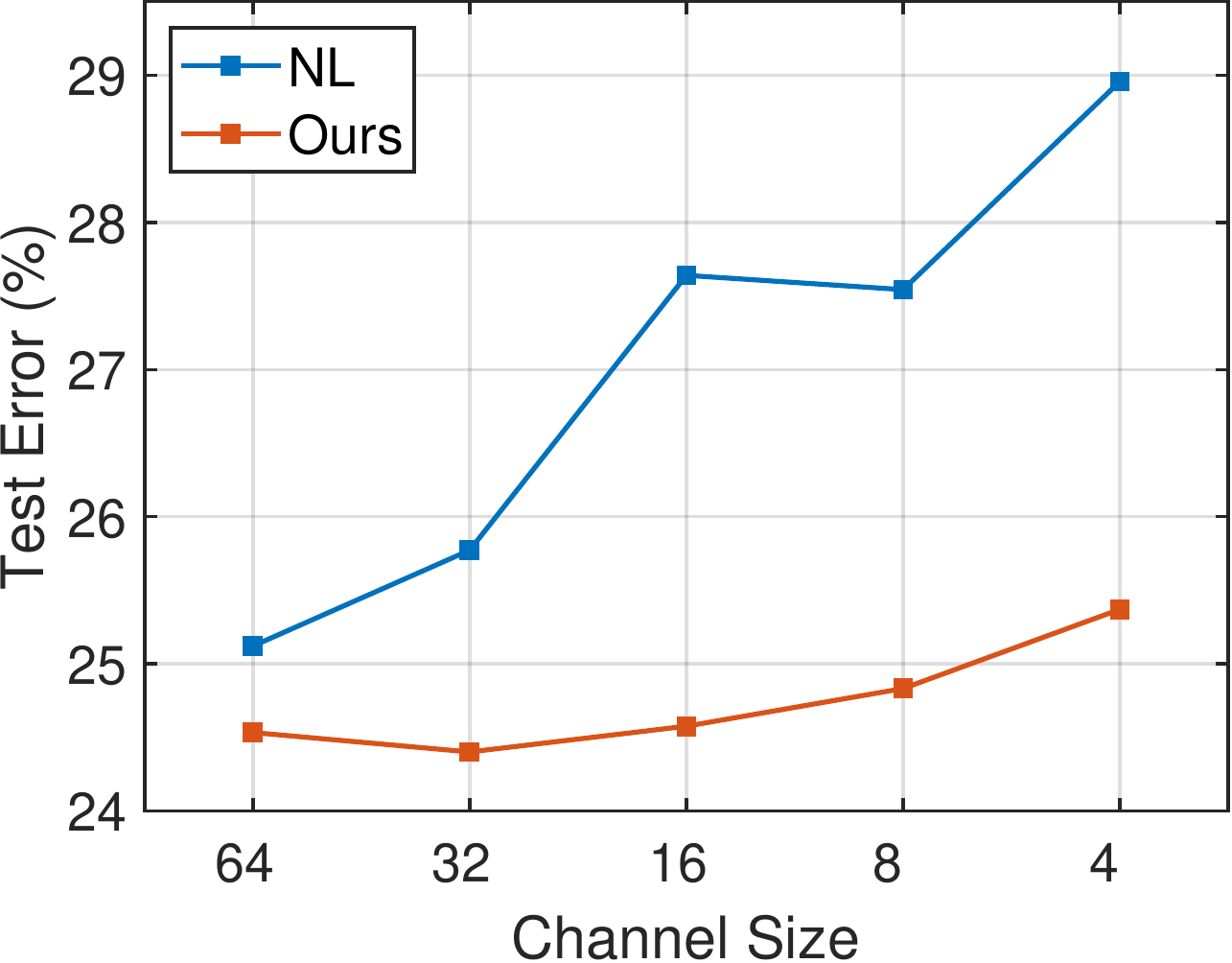}
         \subcaption{CIFAR-100}
    \end{subfigure} 
    \vspace{-3mm}
    \caption{Illustration of test errors(\%) with respect to embedding channel size. We train PreResNet56 by varying the NL blocks on CIFAR-10/100. Both (a) and (b) show that our method improves robustness to embedding channel reduction.}
    \vspace{-4mm}
    \label{fig:ChnnlReduc}
\end{figure}

In this section, we demonstrate the advantages of our method. As confirmed in Section~\ref{subsec:ImpAnls}, attention without \textit{softmax} operation is more likely to learn angular relationships. Hence, we assume our method has the capability to efficiently represent relationships. We verify this by checking the performance as reducing the embedding channel dimension, and obtain the expected results. We train PreResNet56 with 3 NL blocks inserted to the second residual block. As illustrated in Figure~\ref{fig:ChnnlReduc}, our method demonstrates robustness to embedding channel reduction.   

\begin{table}
\begin{center}
\tabcolsep=2mm
\resizebox{0.47\textwidth}{!}{
\begin{tabular}{c|c|c|c|c}
\thickhline
  \multirow{2}{*}{Dataset} & \multirow{2}{*}{Model} & \multicolumn{3}{c}{Initialization} \\ 
  \cline{3-5}
& & \textit{He}~\cite{he2015delving} & $\sigma = 0.01$ & $\sigma = 0.0$  \\
\thickhline
\multirow{2}{*}{CIFAR-10}&PreResNet56+3NL & 5.91 & \textbf{5.73} & 6.23  \\
&PreResNet56+3Ours & \textbf{5.64} & 5.69 & 6.14 \\
\hline
\multirow{2}{*}{CIFAR-100}&PreResNet56+3NL & 25.82 & \textbf{25.12} & 26.24  \\
&PreResNet56+3Ours & \textbf{24.53} & 24.62 & 26.24 \\
\hline

\multirow{2}{*}{Tiny-ImageNet}&PreResNet50+3NL & 35.84 & \textbf{35.35}
 & 37.5  \\
&PreResNet50+3Ours & \textbf{35.08} & 35.34 & 37.23 \\
\thickhline
\end{tabular}}
\end{center}
\vspace{-3mm}
\caption{ Comparison of test errors($\%$) on PreResNet50 and PreResNet56 with CIFAR-10/100 and Tiny-ImageNet. Results of CIFAR and Tiny-ImageNet are averaged over 10 and 3 random seeds, respectively.}
\label{tab:init}
\end{table}

Our method also demonstrates robustness to embedding weight initialization. We compare the performance with three initialization methods: \textit{He initialization}~\cite{he2015delving} and initialization with standard deviation of $\{ 0.0, 0.01\}$. As shown in Table~\ref{tab:init}, the weight of NL block should be initialized with a standard deviation of $0.01$, which is suggested in ~\cite{wang2018non}. It is tuned magic number, and much smaller than the standard deviation of the \textit{He initialization}. By contrast, our method shows the best performance with the standard \textit{He initialization}.

\begin{table*}[t]
    \begin{center}
        \begin{tabular} {c|c|c|c|c|c}
            \thickhline
            \multirow{2}{*}{} & \multirow{2}{*}{Methods} & \multicolumn{4}{c}{Number of heads}\\
            \cline{3-6}
             & & 0&1&2&4\\
            \thickhline
            \multirow{2}{*}{\shortstack{Test Error @ CIFAR-10\\ {[ \% ]} }} & NL & \multirow{2}{*}{6.37}
 & 5.73 & 5.69 & 5.62 \\
            \cline{2-2} \cline{4-6}
              & Ours & 
 & \textbf{5.64} & \textbf{5.48} & \textbf{5.43} \\
            \hdashline
            \multirow{2}{*}{\shortstack{Test Error @ CIFAR-100 \\ {[ \% ]} }} & NL & \multirow{2}{*}{26.59}
 & 25.12 & 24.63 & 24.43 \\
            \cline{2-2} \cline{4-6}
              & Ours & 
 & \textbf{24.53} & \textbf{24.17} & \textbf{24.12} \\
            \hline
            \multirow{2}{*}{\shortstack{Memory \\ {[ MB ]} } } & NL & \multirow{2}{*}{2827}
 & 3789 & 4269 & 5229 \\
            \cline{2-2} \cline{4-6}
             & Ours & 
 & \textbf{3453} & \textbf{3429} & \textbf{3417} \\
            \hline
            \multirow{2}{*}{\shortstack{Train Step Time \\ {[ ms/batch ]} }} & NL & \multirow{2}{*}{62.37}
 & 83.11  & 88.96 & 100.93 \\
            \cline{2-2} \cline{4-6}
             & Ours & 
 & \textbf{77.92} & \textbf{76.97} & \textbf{76.51} \\
            \thickhline
            
        \end{tabular}
    \end{center}
    \vspace{-6mm}
    \caption{Comparison of test errors($\%$), memory(MB), and train step time(ms/batch) on PreResNet56 with CIFAR-10/100. 3 NL blocks are inserted, and results of NL block with 0 head is obtained on PreResNet56 without NL block. Test errors are averaged over 10 random seeds, and train step times are averaged over 300 iterations. }
    \label{table:multihead}
\end{table*}
\vspace{-2mm}

\begin{table}
\begin{center}
\begin{tabular}{c|c|c|c}
\thickhline
  Dataset & Model & NL & Ours  \\
\thickhline
\multirow{3}{*}{CIFAR-10}&PreResNet32 & 6.81 & \textbf{6.65}  \\
&PreResNet56 & 5.66 & \textbf{5.43} \\
&PreResNet110 & 5.29 & \textbf{4.93} \\
\hline
\multirow{4}{*}{CIFAR-100}&PreResNet32 & 29.89 & \textbf{28.84}  \\
&PreResNet56 & 24.33 & \textbf{24.12} \\
&PreResNet110 & 23.28 & \textbf{22.62} \\
&WRN-28-10 & 18.51 & \textbf{18.18} \\
\hline
Tiny-ImageNet & PreResNet50 & 34.76 & \textbf{34.385}  \\
\thickhline
\end{tabular}
\end{center}
\vspace{-4mm}
\caption{ Comparison of test errors($\%$) on PreResNet32, PreResNet50, and PreResNet56 with CIFAR-10/100 and Tiny-ImageNet. 3 NL blocks with 4 heads are inserted. The Results of CIFAR and Tiny-ImageNet are averaged over 10 and 3 random seeds, respectively.}
\vspace{-2mm}
\label{tab:cls}
\end{table}

\section{Experiments}
\label{sec:exp}
\vspace{-2mm}

In this section, we describe details of experiments. We insert 3 NL blocks with 4 heads to the second residual block of PreResNet and Wide-Residual Networks (WRN). To investigate the inefficacy of \textit{softmax} operation, we conduct experiments by varying the NL blocks on CIFAR-10, CIFAR-100, and Tiny-ImageNet.

For CIFAR datasets, we train networks with 50k training images using the standard data augmentation, and evaluate the top-1 errors on 10k test images. We employ SGD with a mini-batch size of 128. Momentum and weight decay are set to 0.9 and 1e-4, respectively. Learning rate is initially set to 0.1, and divided by 10 at 81 and 122 epochs. Training is stopped at 164 epochs.

For Tiny-ImageNet, we train with the 100k training images using the standard data augmentation with 56 pixels cropping, and evaluate the top-1 error on 10k test images. First, we pretrain PreResNet without NL block on Tiny-ImageNet with SGD and mini-batch size of 128. Momentum and weight decay are set to 0.9 and 1e-4, respectively. Learning rate is initially set to 0.1, and divided by 10 every 30 epochs. Training is stopped at 100 epochs. Then, we insert NL blocks to the pretrained networks, and fine-tune. We set the initial learning rate to 0.01, divide it by 10 at 40 epochs, and finish the training at 60 epochs.

As shown in Table~\ref{tab:cls}, we obtain improved performance on PreResNet and WRN with CIFAR-10/100 and Tiny-ImageNet. We constantly get better results regardless of the depth or width of the networks. We get $0.27\%$ and $0.66\%$ accuracy improvement on PreResNet56 with CIFAR-10 and CIFAR-100, respectively. Additionally, as shown in Table~\ref{table:multihead}, our method reduces computational cost by removing \textit{softmax} operation and employing the associative law. As suggested in~\cite{vaswani2017attention}, we set the embedding channel size to $C/N_h$, where $C$ and $N_h$ are the channel size of input features and number of heads, respectively.  Notably, our method can employ multi-head attention without additional computation cost, because the computation cost of our method is complexity of $HW(C/N_h)^2 \times N_h$
\vspace{-2mm}

\section{Conclusion}
\vspace{-2mm}
In this paper, we investigate the way in which attention maps can be calculated. We empirically analyze the inefficacy of \textit{softmax} operation and superiority of \textit{scaled NL block}. We visualize the attention maps and compare the performance of the \textit{magnitude only NL block} and \textit{direction only NL block} to verify that \textit{softmax} operation makes the attention strongly rely on the magnitude of key vectors. By contrast, our method is more efficiently learn angular relationships using the cosine similarity. Our method demonstrates robustness to embedding channel reduction and embedding weight initialization. In addition, our method generally improves the performance with PreResNet and WRN on CIFAR-10/100 and Tiny-ImageNet. Notably, by employing the associative law, the computational cost of our method is largely reduced to a linear function of the number of pixels, and our method can employ multi-head attention without additional cost.

\vspace{4mm}
\noindent \textbf{Acknowledgements} This work was conducted by Center for Applied Research in Artificial Intelligence
(CARAI) grant funded by DAPA and ADD (UD190031RD).

\bibliographystyle{IEEEbib}
\bibliography{refs}

\end{document}